\documentclass[acmsmall]{acmart}

\AtBeginDocument{%
  }

\setcopyright{acmlicensed}
\copyrightyear{2025}
\acmYear{2025}
\acmDOI{XXXXXXX.XXXXXXX}

\acmJournal{JACM}
\acmVolume{37}
\acmNumber{4}
\acmArticle{111}
\acmMonth{8}

\renewcommand\footnotetextcopyrightpermission[1]{}

\begin{document}

\title{Efficient Test-Time Retrieval Augmented Generation}

\author{Hailong Yin}
\affiliation{%
  \institution{Fudan University}
  \country{China}
}

\author{Bin Zhu}
\affiliation{%
 \institution{Singapore Management University}
 \country{Singapore}}

\author{Jingjing Chen}
\affiliation{%
  \institution{Fudan University}
  \country{China}}

\author{Chong-Wah Ngo}
\affiliation{%
 \institution{Singapore Management University}
 \country{Singapore}}


\renewcommand{\shortauthors}{Trovato et al.}

\begin{abstract}
Although Large Language Models (LLMs) demonstrate significant capabilities, their reliance on parametric knowledge often leads to inaccuracies. Retrieval Augmented Generation (RAG) mitigates this by incorporating external knowledge, but these methods may introduce irrelevant retrieved documents, leading to inaccurate responses. While the integration methods filter out incorrect answers from multiple responses, but lack external knowledge like RAG methods, and their high costs require balancing overhead with performance gains. To address these issues, we propose an Efficient Test-Time Retrieval-Augmented Generation Framework named \textbf{ET$^2$RAG} to improve the performance of LLMs while maintaining efficiency. 
Specifically, \textbf{ET$^2$RAG} is a training-free method, that first retrieves the most relevant documents and augments the LLMs to efficiently generate diverse candidate responses by managing response length. Then we compute the similarity of candidate responses and employ a majority voting mechanism to select the most suitable response as the final output. In particular, we discover that partial generation is sufficient to capture the key information necessary for consensus calculation, allowing us to effectively perform majority voting without the need for fully generated responses. Thus, we can reach a balance between computational cost and performance by managing the response length for the number of retrieved documents for majority voting. Experimental results demonstrate that \textbf{ET$^2$RAG} significantly enhances performance across three tasks, including open-domain question answering, recipe generation and image captioning.
\end{abstract}




\maketitle
\makeatletter
\AtBeginDocument{%
  \markboth{}{}
  \pagestyle{empty}
  \thispagestyle{empty}
}
\makeatother
\section{Introduction}
\begin{figure}
  \centering
  \includegraphics[width=\linewidth]{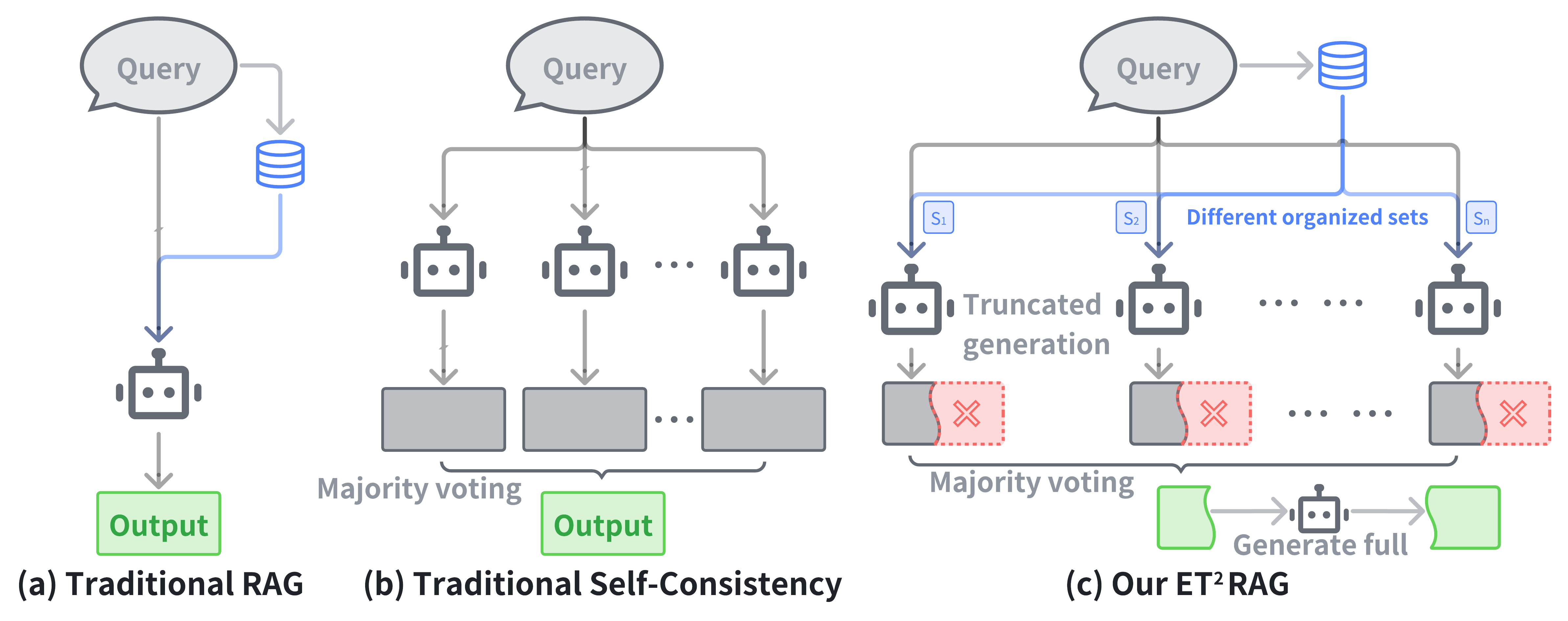}
  \caption{Comparison of retrieval and integration strategies. (a) \textit{Traditional RAG:} Retrieves documents from an external database and feeds them directly into a language model, which may lead to noisy or inconsistent outputs. (b) \textit{Self-Consistency Methods}: Rely on the stochasticity of LLM decoding to generate multiple reasoning paths and select an answer via majority voting, but lack access to external knowledge. (c) \textit{Our ET$^2$RAG }: Integrates retrieval augmentation into self-consistency by first organizing retrieved results into diverse and stable subsets. These subsets are used to generate truncated outputs, which are evaluated via consensus to select the most reliable response. Full generation is performed only once on the best retrieval set, significantly reducing computational overhead while enhancing accuracy and robustness.}
  \label{fig:introim}
\end{figure}
Large Language Models (LLMs) have significantly advanced in recent years, demonstrating remarkable performance across a variety of natural language generation tasks \cite{ouyang2022training, zhao2023survey}. Despite these advancements, LLMs continue to struggle with factual inaccuracies, primarily due to their reliance on parametric knowledge embedded during training, which may become outdated or insufficient for precise and factual responses \cite{mallen2022not,min2023factscore}. To tackle this problem, retrieval-augmented generation (RAG) methods \cite{ram2023context,asai2023retrieval} have been introduced to enhance model outputs by incorporating external knowledge retrieved from relevant databases, as shown in Figure \ref{fig:introim} (a). 
However, traditional RAG methods face two notable limitations. First, the retrieval process frequently includes irrelevant or misleading documents, degrading the accuracy and reliability of the generated responses \cite{shi2023large,gao2023enabling}. Second, only increasing the number of retrieved documents may not improve model performance, and may even lead to worse performance for the LLMs \cite{lewis2020retrieval}.

Recent studies have explored alternative integration strategies, such as multi-agent collaborative frameworks and self-consistency mechanisms, to refine outputs by leveraging consensus across multiple generated responses \cite{wang2022self,wan2024knowledge, du2023improving, wuautogen},
as shown in Figure \ref{fig:introim} (b). 
These methods are capable of selecting the correct answer from multiple candidates by majority voting. For example, LLM-Debate \cite{wuautogen} engages multiple agents in structured discussions to improve reasoning accuracy while CoT-SC \cite{wang2022self} generates multiple reasoning paths to select the most consistent and correct response \cite{wei2022chain}.
Nevertheless, these methods lack external knowledge integration like RAG, limiting performance gains. Additionally, the integration of multiple LLM agents often incurs high costs, necessitating a balance between overhead and performance gains.


Motivated by these insights, we hypothesize that integrating retrieval-augmented generation with consensus-based integration frameworks could simultaneously address the limitations of both approaches. Specifically, such integration could filter out inaccuracies resulting from noisy retrieved information, enhance diversity and robustness of responses, and leverage external knowledge effectively. To this end, 
we propose a novel Efficient Test-Time Retrieval-Augmented Generation (ET$^2$RAG) framework designed to efficiently combine retrieval augmentation with consensus-based integration. Our framework comprises three main steps. First, we introduce Stable Organized Retrieval, which strategically regroups retrieved documents into multiple combinations to ensure robust performance across different retrieval conditions and to mitigate discrepancies in document quality. Second, these combinations are subsequently fed into LLMs to generate multiple candidate responses. Nevertheless, generating the complete response for each combination is computationally expensive. To address this, we introduce partial generation, which extracts key information from each candidate while significantly reducing computational overhead. Finally, we compute similarity scores among all candidate responses and use a majority voting mechanism to identify the response with the highest level of consensus. As a result, ET$^2$RAG manages to maintain computational costs and facilitates rapid consistency checks among responses. 


We validate ET$^2$RAG across diverse tasks, including open-domain question answering, recipe generation and image captioning, demonstrating significant performance improvements relative to state-of-the-art methods. 
Additionally, we extensively investigate the efficiency of ET$^2$RAG regarding two key factors—Vote Size (the number of retrieved sets for voting) and Response Length (the length of the generated responses), to explore the trade-off between performance and computation cost in the ET$^2$RAG framework across different tasks and models. Overall, ET$^2$RAG offers a practical and broadly applicable solution that enhances the reliability and effectiveness of LLM outputs in various generation tasks.


\section{RELATED WORK}
\subsection{Retrieval-Augmented Generation}

Recently, retrieval augmentation has enhanced language models by incorporating external knowledge repositories. Models such as REALM\cite{guu2020retrieval} and RAG\cite{lewis2020retrieval} have showcased the advantages of retrieval-based enhancement. These models retrieve relevant information from external databases to offer additional context for generating precise responses. However, the retrieved passages may include irrelevant information, leading to incorrect outputs. A recent study \cite{asai2023self} introduced a retrieval-augmentation method that requires extensively annotated data, so far effective only in text reasoning. Additionally, studies such as \cite{dong2025decoupling} require architectural modifications and fine-tuning for specific language models, resulting in high implementation costs and limited general applicability. Work like \cite{guo2024lightrag} enhances retrieval by extracting entities with large language models and constructing graph-based indices, but this approach also incurs significant costs for knowledge base construction and maintenance.
Recently, similar studies have been pursued for vision-language tasks\cite{barraco2023little,poppi2024safe,sarto2023positive}. The retrieval-augmented generation in the vision-language domain poses distinct difficulties due to the disparities in modalities and variations in model structures \cite{wei2023uniir}. Preliminary efforts have explored this direction, but they often encounter issues with noise in the retrieved knowledge snippets. Compared to textual augmentation, retrieved passages based on visual content may contain more information that is irrelevant to the image content. Therefore, a training-free, ready-to-use, and widely applicable method that supports multimodal tasks is highly needed.  Our work addresses this issue by introducing RAG while using a training-free integration method to mitigate the noise problem in retrieval.

\subsection{LLM Integration}

CoT-SC\cite{wang2022self} employs a variety of chain-of-thought\cite{wei2022chain} prompts to stimulate diverse reasoning processes within a single LLM and determines the final response through majority voting. Subsequent studies like \cite{li2024more, fu2022complexity,li2023making,cobbe2021training,thoppilan2022lamda,lin2023ask} are extensions of CoT-SC, focusing on text reasoning tasks that rely solely on internal model randomness and do not incorporate external knowledge. In contrast, our method integrates retrieved information for external knowledge augmentation. It has proven effective not only in factual knowledge tasks but also in multimodal contexts. Our approach is versatile, supporting a broad spectrum of models, both visual and textual, independent of their size. Unlike methods that depend on internal randomness, our strategy offers a robust and efficient solution applicable across various tasks and model architectures. Additionally, while many integration techniques, like supervised LLM fusion\cite{wan2024knowledge}, enhance LLM capabilities, they largely depend on supervised learning and heavily annotated datasets.

\section{METHOD}
\label{sec:METHOD}

\subsection{Preliminaries}
The theoretical foundation of our approach builds on the principle of self-consistency in large language models, which posits that correct reasoning paths tend to converge on the same answer, while incorrect ones diverge~\cite{wang2022self, li2024more}.
Formally, given a task query $x$, denote $G(x)$ as the output generated by a language model. When producing a set of candidate responses $\{o_1, o_2, \dots, o_V\}$, self-consistency aims to identify the final output $o_{\text{final}}$ through majority voting:
\[
o_{\text{final}} = \arg\max_{o \in O} \sum_{i=1}^{V} \mathbb{I}(o_i = o),
\]
where $\mathbb{I}(\cdot)$ is the indicator function.

The condition for reliable consensus is met when the correct answer $o^*$ appears more frequently than any alternative $o' \neq o^*$:
\[
\sum_{i=1}^{V} \mathbb{I}(o_i = o^*) > \sum_{i=1}^{V} \mathbb{I}(o_i = o'), \quad \forall o' \neq o^*.
\]

To extend this paradigm to retrieval-augmented generation, we incorporate multiple retrieved evidence sets into the generation process. Specifically, each candidate output is conditioned on a distinct retrieved subset $s_i$, such that:
\[
o_i = G(x, s_i),
\]
and the final output is selected via:
\[
o_{\text{final}} = \arg\max_{o \in O} \sum_{i=1}^{V} \mathbb{I}(G(x, s_i) = o).
\]

\begin{figure*}[htbp]
  \centering
  \includegraphics[width=\textwidth]{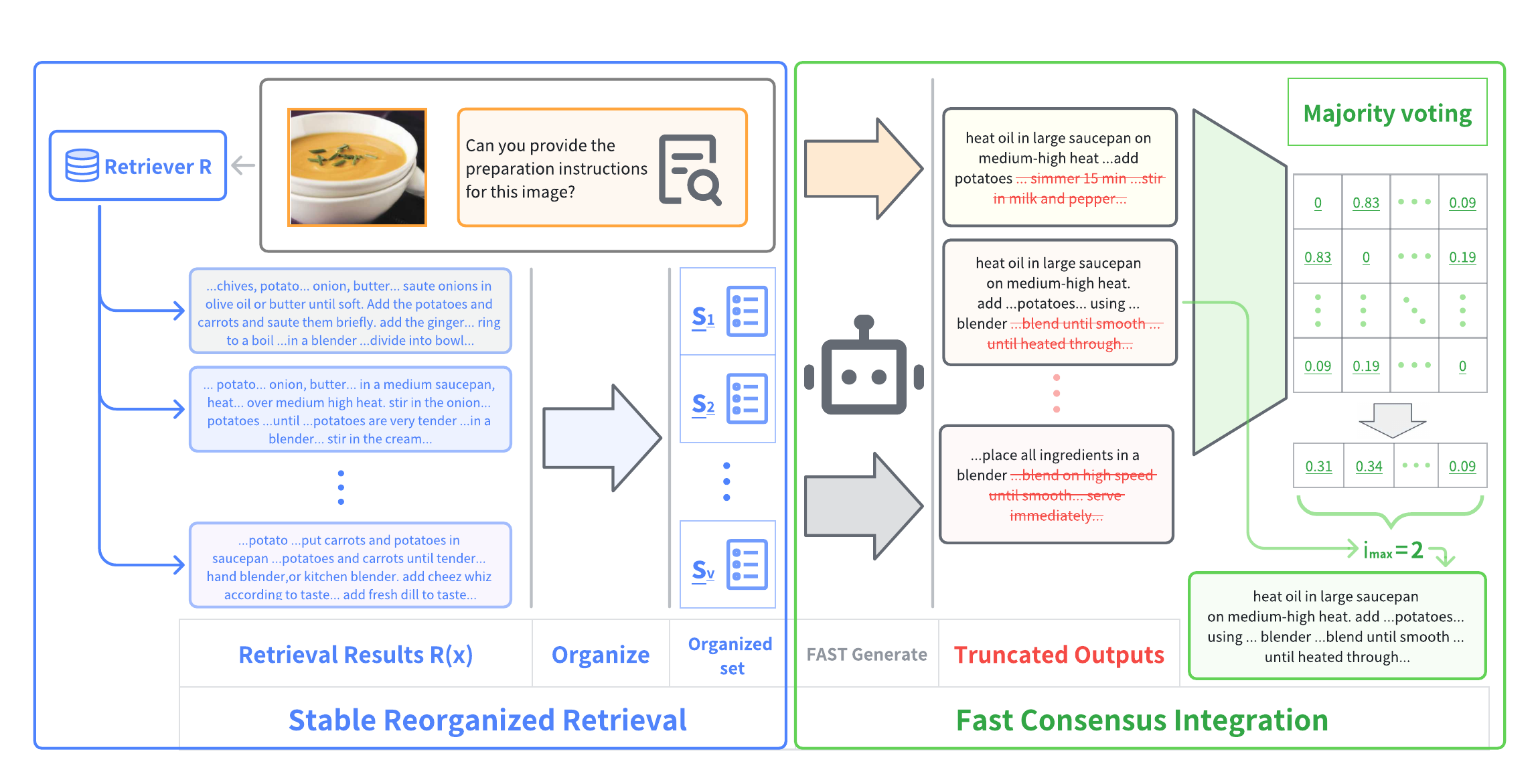}
  \caption{Our proposed ET$^2$RAG consists of two stages: \textbf{Stable Organized Retrieval} and \textbf{Fast Consensus Integration}. In the former stage, the input task query (either text or image) is passed to the \textbf{Retriever}, which retrieves multiple relevant results from the database (\( R(x) \)). These retrieved results are then organized into independent combinations in the organized set (\( S \)) through the \textbf{Organization} process. These combinations form the basis for the subsequent generation phase. In the latter stage, the organized combinations are concatenated with the task query and input into the \textbf{Fast Generation} module to generate truncated outputs of fixed length. Next, \textbf{Consensus Negotiation} is applied to calculate the similarity between these outputs, resulting in a similarity matrix (\( M \)). By summing the elements of this matrix, we compute the \textbf{Agreement Scores} (\( A \)) for each output. Finally, the output corresponding to the highest agreement score is selected, and its associated combination is used to generate the final complete output.}
  \label{fig:methodpic}
\end{figure*}
In this context, the condition for self-consistency hinges on the quality of the retrieved subsets $s_i$. To satisfy the theoretical guarantees, a majority of these subsets must support the correct answer $o^*$, while irrelevant or noisy subsets should result in divergent responses. Thus, ensuring retrieval robustness is key to enabling reliable consensus in the retrieval-augmented setting.

This formulation motivates the design of ET$^2$RAG, which introduces a structured method for organizing retrieved documents and leveraging fast generation to achieve high consensus with minimal computational overhead. We expand on this in the following.



\subsection{Framework Overview}
We propose Efficient Test-Time Retrieval-Augmented Generation (ET$^2$RAG), as depicted in Figure \ref{fig:methodpic}, which consists of two primary stages: Stable Organized Retrieval and Fast Consensus Integration. The first stage, Stable Organized Retrieval, includes the steps of Retrieval and Organization. The Retrieval step systematically extracts relevant information, while the Organization step structures this retrieved information to ensure consistency and relevance. Based on the organized information from the first stage, the second stage, Fast Consensus Integration, comprises Fast Generation, Consensus Negotiation, and Majority Voting. Initially, candidate responses are rapidly generated through partial generation. These responses then undergo a consensus negotiation process, culminating in majority voting to integrate the most reliable and effective results. 


\subsection{Stable Organized Retrieval}
\label{sec:SOR}

\noindent \textbf{Retrieval}. Given a task query $x$, we utilize a retriever $R$ to search a relevant external database $D$ and retrieve a set of documents
\( R(x) = \{ r_1, r_2, \dots, r_N \} \). Each document \( r_i \) contains knowledge potentially beneficial for addressing the task query. These retrieved documents serve as the basis for the subsequent organization process.

\noindent \textbf{Organization}. Due to variability in quality among retrieved documents, we propose an organizational strategy \( \mathcal{T} \) to systematically regroup the documents into multiple balanced subsets. Specifically, we construct an organized set 
$$
S = \mathcal{T}(R(x)) = \{s_1, s_2, \dots, s_V\}
$$
Each subset $s_i$ comprises strategically selected document combinations and $V$ denotes the size of the organized set. 
$V$ also represents Vote Size and the number of candidate outputs that will be generated. This step ensures that each subset provides robust and balanced support for generating candidate responses, significantly mitigating discrepancies in document quality and enhancing overall retrieval consistency. Please see more details of the organization strategy for different tasks in Section~\ref{sec:Experimental_setup}.


\subsection{Fast Consensus Integration}
Fast Consensus Integration aims to efficiently select the most reliable answer from multiple candidates through fast generation, consensus negotiation, and majority voting, to improve both performance and efficiency. 

\noindent \textbf{Fast Generation}.
After obtaining the organized set \( S \) from Section~\ref{sec:SOR}, the reasoning path is jointly driven by the retrieval information and the generative model. We concatenate each combination \( s_i \) with the task query \( x \), forming input pairs \( (x, s_i) \). These pairs are then processed by the generator \( G \), which produces the corresponding outputs \( o_i \):

\[
o_i = G(x, s_i).
\]

To improve efficiency and control computational overhead, we truncate each generated output \( o_i \) to a maximum length \( L \) and only get the partial generation results, ensuring that the output does not exceed this token limit:

\[
o_i = \text{truncate}(G(x, s_i), L).
\]

The truncation strategically reduces computation cost while maintaining the quality of the generated outputs. 
By limiting the output length, we strike a balance between performance and resource consumption, particularly when incorporating retrieval-augmented generation (RAG) into consensus-based integration methods. In practice, we find that truncated outputs can retain sufficient context to evaluate consensus effectively among generated responses and yield high-quality results.
Finally, the collection of all generated outputs is:
$$
O = \{o_1, o_2, \dots, o_V\}.
$$

\noindent \textbf{Consensus Negotiation}.
After obtaining the truncated output set $O$ from the fast generation stage, we further introduce consensus negotiation to compute the agreement among the outputs. Specifically, we use similarity calculation methods $C$ to compute the similarity between each pair of elements in the output set.
Then we can obtain a similarity matrix $M$ of size $V \times V$, where each element $M_{ij}$ represents the consensus agreement between outputs $o_i$ and $o_j$. The matrix elements are defined as follows:
$$
M_{ij} = C(o_i, o_j).
$$
The similarity matrix $M$ captures the degree of agreement among candidate responses, providing a quantitative foundation for selecting the most reliable answer next.

\noindent \textbf{Majority Voting}.
Based on the similarity matrix $M$ obtained in the consensus negotiation, we sum the elements along the natural dimension of the matrix to obtain an array $A$, which represents the overall agreement score of each output with the other outputs. The calculation is given by:
$$
A_i = \sum_{j=1}^{V} M_{ij}, \quad \text{for each } i = 1, 2, \dots, V.
$$


We then identify the index $i_{\text{max}}$ corresponding to the output with the highest agreement score from the array $A$. Finally, we select the corresponding combination $s_{i_{\text{max}}}$ from the organized set $S$ and its truncated output $o_{i_{\text{max}}}$ as previous context based on the chosen index $i_{\text{max}}$ to produce the final output as follows:
$$
i_{\text{max}} = \arg\max_{i \in \{1, 2, \dots, V\}} A_i,
$$
$$
o_{\text{final}} = G(x, s_{i_{\text{max}}}).
$$

Through this majority voting process, we are able to select the most consistent and reliable answer from the candidate results.


The advantage of this process is that it effectively extracts high-quality knowledge combinations from the external database, providing strong support for LLMs to address the task query $x$, thereby enhancing the accuracy and consistency of the generated answers.
As a result, we efficiently leverage retrieval-augmented knowledge without complex processes or heavy computation, enabling fast and reliable responses.

\section{EXPERIMENTS}
\label{sec:EXPERIMENTS}

\subsection{Experimental Setup}
\label{sec:Experimental_setup}

\begin{table}[htbp]
  \caption{Overview of datasets, retrievers, and generators.}
  \label{tab:experiments_overview}
  \begin{tabular}{|c|c|l|}
    \hline
    \textbf{Datasets} & \textbf{Retriever} & \textbf{Generator} \\
    \hline
    {TriviaQA\cite{joshi2017triviaqa}} & {Contriever-MS\cite{izacard2022unsuperviseddenseinformationretrieval}} & Llama2$_{7B}$\cite{touvron2023llama} \\
     &  & Llama3$_{8B}$\cite{grattafiori2024llama} \\
     &  & DeepSeek-R1-Llama$_{8B}$\cite{guo2025deepseek} \\
    \hline
    {PopQA\cite{mallen2022not}} & {Contriever-MS\cite{izacard2022unsuperviseddenseinformationretrieval}} & Llama2$_{7B}$\cite{touvron2023llama} \\
     &  & Llama3$_{8B}$\cite{grattafiori2024llama} \\
     &  & DeepSeek-R1-Llama$_{8B}$\cite{guo2025deepseek} \\
    \hline
    Recipe1M\cite{salvador2017learning} & Recipe retrieval model\cite{liu2024retrievalaugmentedrecipegeneration} & LLaVA-FT-RAG\cite{liu2024retrievalaugmentedrecipegeneration} \\
    \hline
    COCOTest\cite{chen2015microsoftcococaptionsdata} & CLIP-ResNet-50x64\cite{radford2021learningtransferablevisualmodels} & SMALLCAP\cite{ramos2023smallcap} \\
    \hline
  \end{tabular}
\end{table}

We evaluate our proposed ET$^2$RAG framework across various tasks and datasets.  In line with previous works~\cite{asai2023self, liu2024retrievalaugmentedrecipegeneration, ramos2023smallcap}, the datasets, retrievers, and generators used in the experiments are summarized in Table \ref{tab:experiments_overview}.

\noindent \textbf{Tasks and Datasets.} Our method is assessed on three representative tasks:
(1) \textbf{Open-domain Question Answering tasks}. We conduct experiments on PopQA~\cite{mallen2022not} and TriviaQA\cite{joshi2017triviaqa}, which focus on answering factual knowledge questions. For PopQA, we use long-tail subset as our test set, following~\cite{asai2023self}, and for TriviaQA, we use the test split in~\cite{min2019discrete, guu2020retrieval} for evaluation. Performance is measured in terms of Accuracy (acc). Following \cite{mallen2022not, schick2024toolformer,asai2023self}, the results are evaluated by checking whether the generated answers include the gold answer or not, rather than the exact match.
(2) \textbf{Recipe Generation} is a vision-language task that focuses on generating detailed and coherent cooking instructions based on input food images \cite{chhikara2024fire, yin2023foodlmm, liu2024retrievalaugmentedrecipegeneration}. We use the Recipe1M dataset \cite{salvador2017learning} for experiments, and use the metrics BLEU, SacreBLEU, and Rouge-L for evaluation, in line with prior work \cite{liu2024retrievalaugmentedrecipegeneration}.
(3) \textbf{Image Captioning} is tested on the COCO dataset~\cite{chen2015microsoft}, using the standard Karpathy splits~\cite{karpathy2015deep}. Unlike earlier work~\cite{ramos2023smallcap}, which evaluated a random sample of 1,000 data points from the test set, we evaluate the entire test set. We also compute the metrics: BLEU-4, METEOR, CIDEr for evaluation.
\begin{table}[htbp]
\caption{Details of Organization Strategy.}
\centering
\begin{tabular}{|c|c|c|c|}
\hline
\textbf{TriviaQA} & \textbf{PopQA} & \textbf{Recipe1M} & \textbf{CoCO} \\
\hline
$\text{top}_1$ & $\text{top}_{1}$ & $\text{top}_1$ & $\text{top}_{1,2,3,4}$ \\
$\text{top}_{1,2}$ & $\text{top}_{1,2}$ & $\text{top}_2$ & $\text{top}_{1,2,5,6}$ \\
$\text{top}_{1,3}$ & $\text{top}_{1,3}$ & $\text{top}_3$ & $\text{top}_{3,4,5,6}$ \\
$\text{top}_{1,4}$ & $\text{top}_{1,4}$ & $\text{top}_4$ & $\text{top}_{1,2,7,8}$ \\
$\text{top}_{1,5}$ & $\text{top}_{1,5}$ & $\text{top}_5$ & $\text{top}_{3,4,7,8}$ \\
$\text{top}_{1,6}$ & $\text{top}_{1,6}$ & $\text{top}_6$ & $\text{top}_{5,6,7,8}$ \\
$\text{top}_{1,7}$ & $\text{top}_{1,7}$ & $\text{top}_7$ & $\text{top}_{1,2,9,10}$ \\
$\text{top}_{1,8}$ & $\text{top}_{1,8}$ & $\text{top}_8$ & $\text{top}_{3,4,9,10}$ \\
$\text{top}_{1,9}$ & $\text{top}_{1,9}$ & $\text{top}_9$ & $\text{top}_{5,6,9,10}$ \\
...... & ...... & ...... & ...... \\
\hline
\end{tabular}
\label{tab:detailsre}
\end{table}
\noindent \textbf{Retriever and Generator.} Following previous works~\cite{asai2023self, liu2024retrievalaugmentedrecipegeneration, ramos2023smallcap}, we utilize Contriever-MS\cite{izacard2021unsupervised} as the retriever for open-domain question answering task, image-to-recipe retrieval model\cite{salvador2021revamping} for recipe generation and  CLIP-ResNet-50x64~\cite{radford2021learningtransferablevisualmodels} for image captioning. We use the Llama2$_{7B}$\cite{touvron2023llama} for open-domain question answering. To better demonstrate the generality of our method, we also conduct additional experiments on two other large language models (LLMs): \textit{DeepSeek-R1-Distill-Llama-8B} (referred to as \textit{DeepSeek-Llama\textsubscript{8B}}) and \textit{Meta-Llama-3-8B} (referred to as \textit{Llama3\textsubscript{8B}}).
For recipe generation, we employ LLaVA-Finetuned (LLaVA-FT) and the best-trained LLaVA-Finetuned with retrieval augmentation (LLaVA-FT-RAG) from \cite{liu2024retrievalaugmentedrecipegeneration}. For image captioning, we follow the best practices from \cite{ramos2023smallcap} with the SMALLCAP model.


\noindent \textbf{Organization}.
We apply different organization strategies tailored to task-specific properties, as shown in Table~\ref{tab:detailsre}.
First, open-domain question answering deals with pure text, where the quality of documents in higher-ranked positions significantly surpasses that of lower-ranked ones in text retrieval results. Therefore, for both datasets, we adopt a combination of $\{\text{top}_1, \text{top}_k\}$ to balance retrieval quality and the integration of relevant information from lower-ranked documents as shown in the PopQA column and TriviaQA column of Table~\ref{tab:detailsre}.
Second, in the context of recipe generation, cross-modal recipe retrieval presents significant challenges, primarily due to the complexities of cause-and-effect relationships and the modeling of extended texts~\cite{salvador2017learning, zhu2019r2gan, song2024enhancing}. This often results in the majority of retrieval outcomes having relatively similar quality levels. Consequently, each retrieval result is typically capable of providing pertinent information. Additionally, the length of token sequences in recipe retrieval results can be excessive for the model. Prior work trained models using only a single retrieval result; therefore, we maintain consistency by organizing the retrieved $\{\text{top}_k\}$ recipes as context, referencing the Recip1M column of Table~\ref{tab:detailsre}.
Third, in the domain of image captioning, adhering to the best practices of prior work, we maintain the same number of retrieved captions as reported in \cite{ramos2023smallcap}. From the top 20 retrieved results, we select four captions as a group to form various combinations. Our organization strategy utilizes a simple permutation and combination approach, demonstrated in the CoCo column of Table~\ref{tab:detailsre}, aiming to balance retrieval quality and diversity while ensuring that the retrieved information is both comprehensive and consistent in quality.

\subsection{Performance Comparison}

Table \ref{tab:OQA_result}, Table \ref{tab:RCP_result} and Table \ref{tab:smallcap_result} list the performance comparison with baselines and traditional RAG method for open domain question answering, recipe generation and image captioning respectively.

\begin{table}[htbp]
  \caption{Performance of Open-domain Question Answering.}
  \label{tab:OQA_result}
  \begin{tabular}{|l|c|c|}
    \hline
    \textbf{Method} & \textbf{PopQA (acc)} & \textbf{TriviaQA (acc)} \\
    \hline
    Llama2$_{7B}$\cite{touvron2023llama} & 14.7\% & 30.5\% \\
    RAG(Llama2$_{7B}$\cite{touvron2023llama}) & 38.2\% & 42.5\% \\
    \textbf{ET$^2$RAG}(Llama2$_{7B}$\cite{touvron2023llama}) & \textbf{44.7\%} & \textbf{52.6\%} \\
    \hline 
    Llama3$_{8B}$ & 15.2\% & 34.5\% \\
    RAG(Llama3$_{8B}$) & 39.5\% & 43.4\% \\
    \textbf{ET$^2$RAG}(Llama3$_{8B}$) & \textbf{51.5\%} & \textbf{59.0\%} \\
    \hline 
    DeepSeek-R1-Llama$_{8B}$ & 15.5\% & 36.7\% \\
    RAG(DeepSeek-R1-Llama$_{8B}$) & 42.5\% & 50.7\% \\
    \textbf{ET$^2$RAG}(DeepSeek-R1-Llama$_{8B}$) & \textbf{52.0\%} & \textbf{60.5\%} \\
    \hline
  \end{tabular}
\end{table}

\begin{table}[htbp]
  \caption{Performance Comparison of Recipe Generation.}
  \label{tab:RCP_result}
  \begin{tabular}{|l|c|c|c|}
    \hline
    \textbf{Method} & \textbf{BLEU} & \textbf{SacreBLEU} & \textbf{ROUGE L} \\
    \hline
    FoodLMM\cite{yin2023foodlmm} & 27.86 & 6.24 & 36.96 \\
    \hline
    LLaVA-FT\cite{liu2024retrievalaugmentedrecipegeneration} & 28.32 & 5.88 & 38.18 \\
    \hline
    RAG(LLaVA-FT-RAG\cite{liu2024retrievalaugmentedrecipegeneration}) & 29.23 & 6.21 & 38.43 \\
    \hline
    \textbf{ET$^2$RAG}(LLaVA-FT-RAG\cite{liu2024retrievalaugmentedrecipegeneration}) & \textbf{30.10} & \textbf{6.45} & \textbf{39.12} \\
    \hline
  \end{tabular}
\end{table}

\begin{table}[htbp]
  \caption{Performance Comparison of Image Captioning.}
  \label{tab:smallcap_result}
  \begin{tabular}{|l|c|c|c|}
    \hline
    \textbf{Method} & \textbf{BLEU-4} & \textbf{METEOR} & \textbf{CIDEr} \\
    \hline
    RAG(SMALLCAP\cite{ramos2023smallcap}) & 36.58 & 27.70 & 118.03 \\
    \hline
    \textbf{ET$^2$RAG}(SMALLCAP\cite{ramos2023smallcap}) & \textbf{37.09} & \textbf{27.99} & \textbf{120.15} \\
    \hline
  \end{tabular}
\end{table}
For \textbf{Open-Domain Question Answering} (Table \ref{tab:OQA_result}), ET\textsuperscript{2}RAG demonstrates significant performance improvements across all tasks and models compared to their respective baselines and the standard retrieval-augmented generation (RAG) method. Specifically, on the \textbf{PopQA} dataset, ET\textsuperscript{2}RAG applied to the Llama2$_{7B}$ model improves its performance by \textbf{+30.0\%} (from 14.7\% to 44.7\%) and outperforms the standard RAG method by \textbf{+6.5\%} (from 38.2\% to 44.7\%). Similarly, for the Llama3$_{8B}$ model, ET\textsuperscript{2}RAG achieves a \textbf{+36.3\%} improvement (from 15.2\% to 51.5\%) and surpasses the RAG method by \textbf{+12.0\%} (from 39.5\% to 51.5\%). Furthermore, when applied to the DeepSeek-R1-Llama$_{8B}$ model, ET\textsuperscript{2}RAG improves performance by \textbf{+36.5\%} (from 15.5\% to 52.0\%) and outperforms the RAG method by \textbf{+9.5\%} (from 42.5\% to 52.0\%).On the \textbf{TriviaQA} dataset, ET\textsuperscript{2}RAG also shows notable enhancements. For the Llama2$_{7B}$ model, it improves performance by \textbf{+22.1\%} (from 30.5\% to 52.6\%) and surpasses the RAG method by \textbf{+10.1\%} (from 42.5\% to 52.6\%). For the Llama3$_{8B}$ model, ET\textsuperscript{2}RAG achieves a \textbf{+24.5\%} improvement (from 34.5\% to 59.0\%) and outperforms the RAG method by \textbf{+15.6\%} (from 43.4\% to 59.0\%). Finally, for the DeepSeek-R1-Llama$_{8B}$ model, ET\textsuperscript{2}RAG improves performance by \textbf{+23.8\%} (from 36.7\% to 60.5\%) and surpasses the RAG method by \textbf{+9.8\%} (from 50.7\% to 60.5\%).

For \textbf{Recipe Generation} (Table \ref{tab:RCP_result}), ET\textsuperscript{2}RAG achieves superior performance compared to the baseline models and the standard RAG method which retrieves the $\text{top}_1$ recipe. When applied to the LLaVA-FT-RAG model, ET\textsuperscript{2}RAG improves BLEU by \textbf{+0.87} (from 29.23 to 30.10), SacreBLEU by \textbf{+0.24} (from 6.21 to 6.45), and ROUGE L by \textbf{+0.69} (from 38.43 to 39.12). Additionally, compared to the FoodLMM baseline, our method improves BLEU by \textbf{+2.24} (from 27.86 to 30.10), SacreBLEU by \textbf{+0.21} (from 6.24 to 6.45), and ROUGE L by \textbf{+2.16} (from 36.96 to 39.12).

For \textbf{Image Captioning} (Table \ref{tab:smallcap_result}), ET\textsuperscript{2}RAG demonstrates consistent improvements over the standard RAG method. Specifically, it improves BLEU-4 by \textbf{+0.51} (from 36.58 to 37.09), METEOR by \textbf{+0.29} (from 27.70 to 27.99), and CIDEr by \textbf{+2.12} (from 118.03 to 120.15) when applied to the SMALLCAP model. These results validate the effectiveness and generalizability of ET\textsuperscript{2}RAG across diverse tasks.

\begin{figure*}[htbp]
  \centering
    \includegraphics[width=\textwidth]{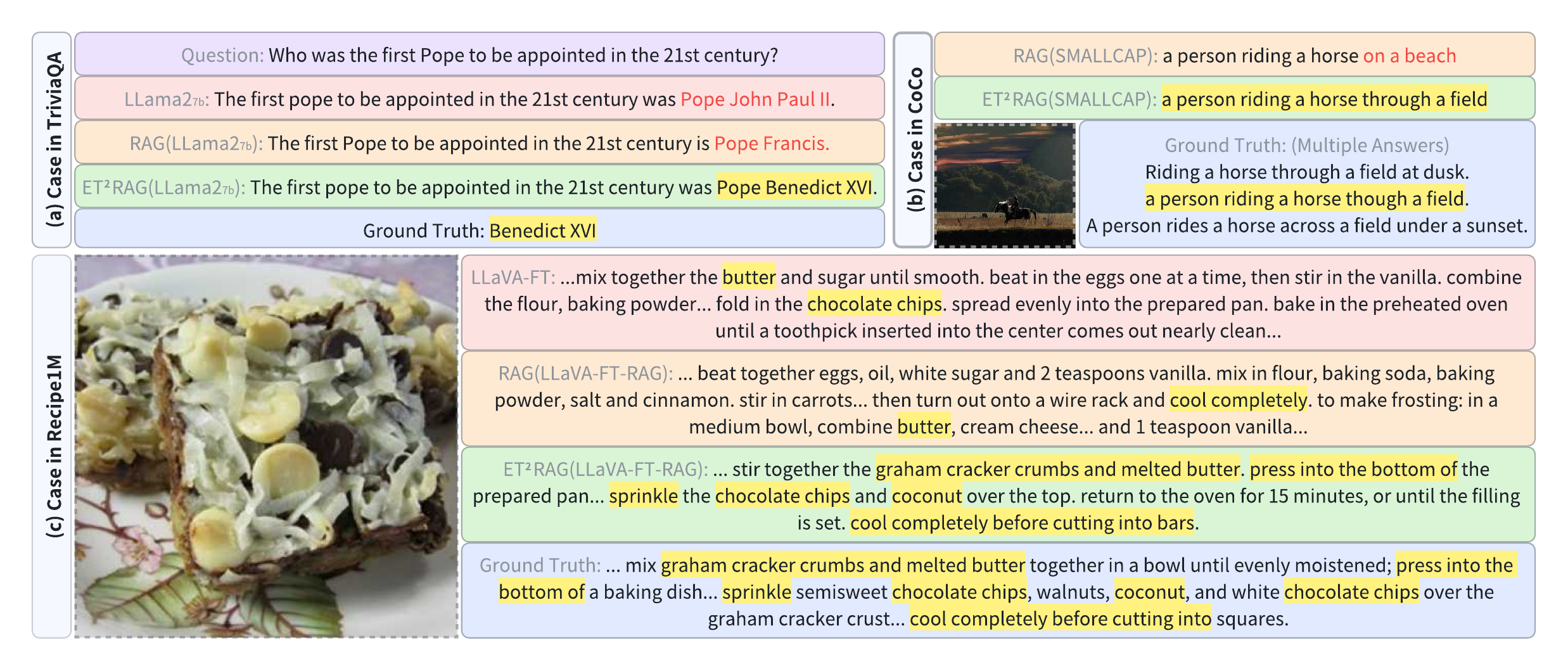} 
    \caption{Qualitative results are presented, with responses matching the ground truth in yellow and incorrect outputs in red.}

  \label{fig:showcase}
\end{figure*}
As shown in the qualitative examples in Figure \ref{fig:showcase}, ET$^2$RAG consistently outperforms existing methods, providing more accurate and relevant outputs. For instance, in Figure \ref{fig:showcase} (a) ET$^2$RAG(LLama2$_{7b}$) correctly identifies Pope Benedict XVI as the first pope in the 21st century, while LLama2$_{7b}$ and RAG(LLama$_{7b}$) fail. In addition, ET$^2$RAG(LLaVA-FT-RAG) generates the correct dessert recipe for the given food image, while LLaVA-FT and RAG(LLaVA-FT-RAG) produce irrelevant cake or carrot cake recipes in Figure \ref{fig:showcase} (c).

\subsection{Ablation Study}
\begin{figure*}[htbp]
  \centering
    \includegraphics[width=\textwidth]{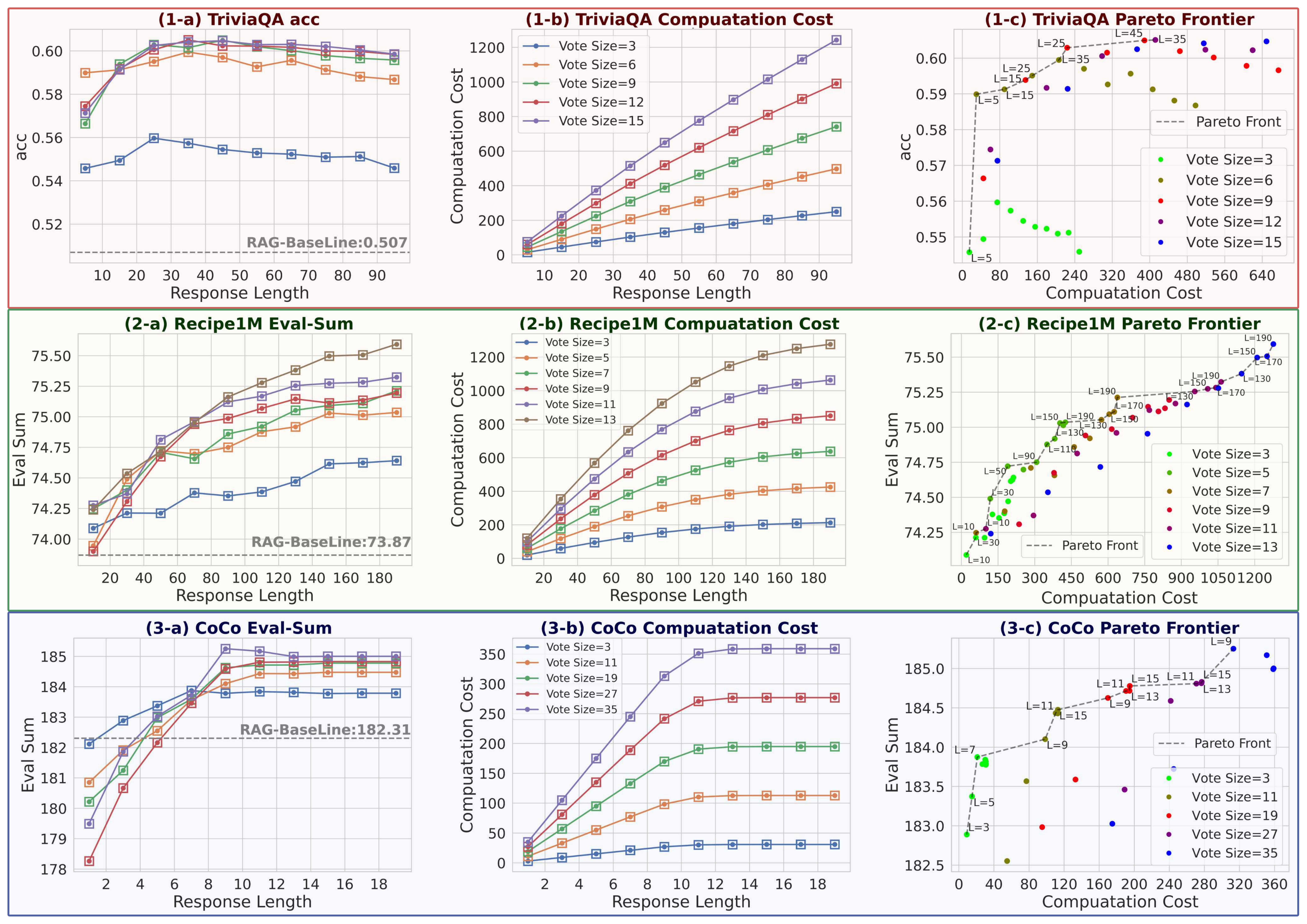} 

    \caption{Results for TriviaQA, Recipe1M, and CoCo. Figure 1-a illustrates the (\textit{acc}) of \textbf{ET$^2$RAG} (DeepSeek-R1-Llama$_{8B}$) for TriviaQA. Figure 2-a shows the `\textit{Eval Sum}', representing the combined scores of BLEU, SacreBLEU, and Rouge-L for Recipe1M. Figure 3-a displays the `\textit{Eval Sum}' for CoCo Captioning, calculated as the sum of BLEU-4, METEOR, and CIDEr scores. The RAG baseline represents results from the traditional RAG method. Details of the metrics are in supplementary material Section 2, A. Figures 1-b, 2-b, and 3-b detail the `\textit{Computation Cost}', capturing the average number of additional tokens produced per response with varying $L$ and $V$ in each task. Figures 1-c, 2-c, and 3-c depict the respective Pareto Frontiers for each task. These frontiers optimize the key metrics (`\textit{Eval Sum}' or accuracy) and `\textit{Computation Cost}' by the strategic adjustment of parameters $L$ and $V$. Each task's optimization strategy equally prioritizes the maximization of its specific evaluation metrics and the minimization of computation costs. \textit{Note} that Figures 1-a, 2-a, and 3-a share legends with Figures 1-b, 2-b, and 3-b, respectively.}

  \label{fig:ALL}
\end{figure*}

\begin{figure}[htbp]
  \centering

    \includegraphics[width=\columnwidth]{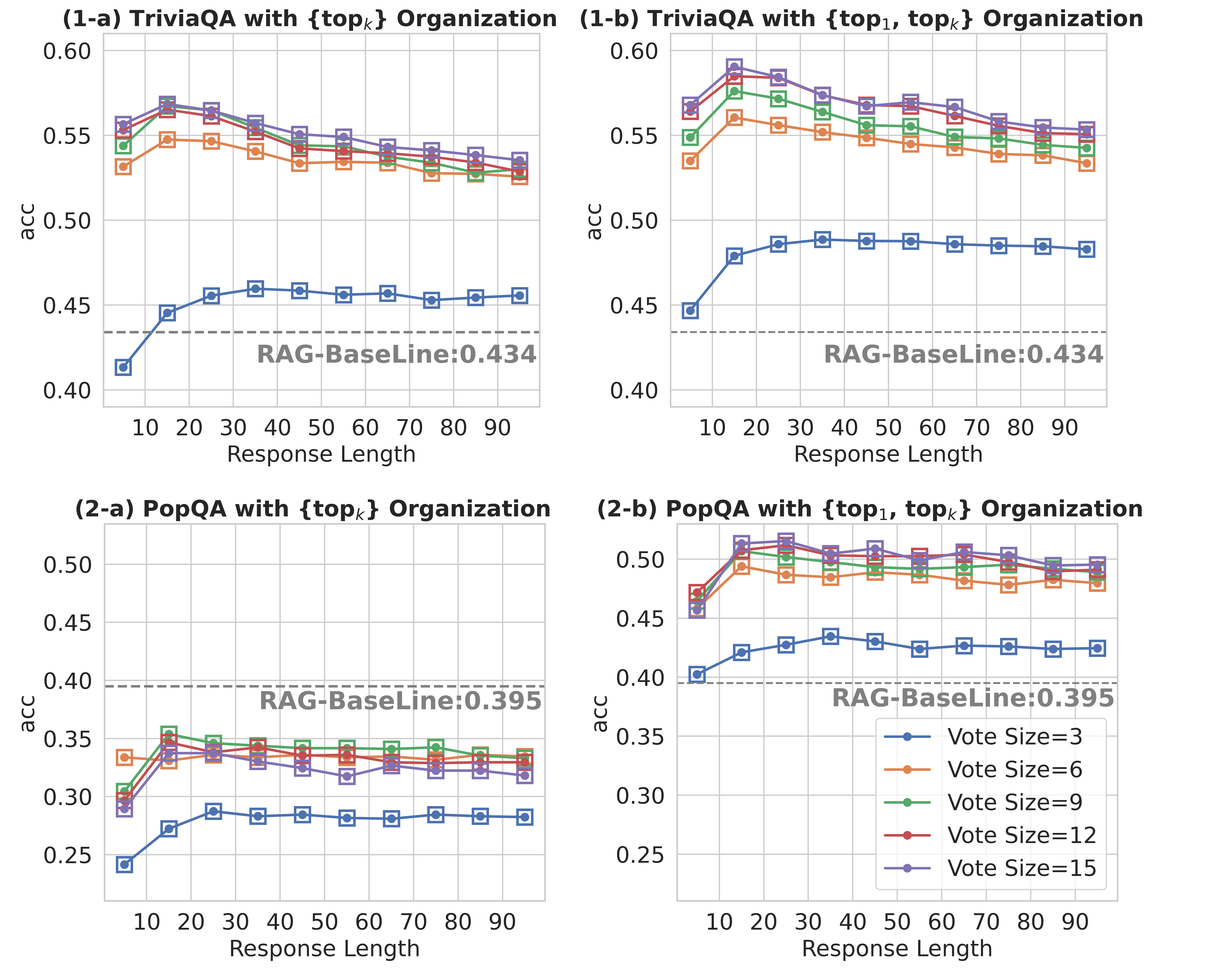} 
    \caption{Discussion for Organization Strategy in PopQA and TriviaQA(Reported when using  \textbf{ET$^2$RAG}(Llama3$_{8B}$). The RAG baseline represents results from the traditional RAG.}
  \label{fig:abl}
\end{figure}

\noindent \textbf{Response Length ($L$) and Vote Size ($V$)}.
As shown in Figure \ref{fig:ALL} (1-a), Figure \ref{fig:ALL} (2-a) and Figure \ref{fig:ALL} (3-a), the performances vary with different $L$ and $V$. Interestingly, the experimental results expose different trends in tasks.
Specifically, in the TriviaQA dataset, when using DeepSeek-R1-Llama$_{8B}$, 
depicted in Figure \ref{fig:ALL} (1-a), the accuracy reaches peak when $L$ is relatively small but with a sufficiently augmented $V$. Its subsequent decrease correlates with increases in $L$. This phenomenon results from correct answers typically emerging when the output generated by the model is shorter (that is, when \( L < 50 \)). When retrieval results provide sufficient support for the question, the model often directly presents the answer at the beginning of the response. This is typically the optimal moment for Majority Voting. In contrast, subsequent responses tend to introduce more noise as they attempt to explain causal relationships, which diminishes the effectiveness of Majority Voting. As the response length increases, the model begins to describe causal relationships rather than focusing on the direct answer to the question, which leads to a more ambiguous consensus among the multiple outputs. This, in turn, results in an inaccurate Majority Voting outcome. Detailed case can be found in the supplementary material Section 3. 
In particular, as shown in Figure \ref{fig:ALL} (1-a), 
the accuracy of the TriviaQA datasets exhibits a upward trend at first with the increase of $V$. However, when $V$ exceeds a certain threshold, the accuracy tends to saturate, showing marginal stabilization with further increments in $V$. This saturation phenomenon can be attributed to the degraded quality of data in the late combination of the organized set $S$. A similar phenomenon can also be observed in PopQA. The Llama3$_{8B}$ model exhibits nearly identical behavior, as illustrated in Figure \ref{fig:abl}.

Additionally, for recipe generation, depicted in Figure \ref{fig:ALL} (2-a), an ongoing improvement trend is observed with the increase of $L$, though the pace of such an increase decelerates and eventually levels off. This continued improvement is attributed to the fact that the recipe is generally long document with hundreds of words and a wide range of retrieved recipes are capable of providing relevant information such as ingredients and cooking methods.
Correspondingly, the image captioning in COCO dataset exhibited a similar trend as represented in Figure \ref{fig:ALL} (3-a). The performance is generally improved with increasing $V$ with a peak in $V$=35 and the best response length is $L$=9.

\subsection{Efficiency Analysis}
\label{sec:efficiency}
To measure the overhead clearly, we define the ``Computation Cost`` to be the average number of extra tokens generated per response as the additional cost introduced by the framework. The Computation Cost for the TriviaQA, Recipe1M and CoCo are shown in Figure \ref{fig:ALL} (1-b), Figure \ref{fig:ALL} (2-b) and Figure \ref{fig:ALL} (3-b).

To further investigate the efficiency of our ET$^2$RAG and analyze the balance between performance and computational cost, we adopt the concept of Pareto Optimality\cite{ngatchou2005pareto}. Our goal is to identify configurations that achieve an optimal balance between two competing objectives:
    (1) \textbf{Objective 1:} Minimize the Computation Cost introduced by the ET$^2$RAG framework.
    (2) \textbf{Objective 2:} Maximize the task-specific performance metrics under the ET$^2$RAG framework.
These two objectives represent a typical multi-objective optimization problem~\cite{deb2011multi}.
Thus, we treat these objectives and the hyperparameters $L$ and $V$ as the search space for optimization. A grid search is conducted over values of $L$ and $V$, as shown in Figure \ref{fig:ALL} (1-c), Figure \ref{fig:ALL} (2-c) and Figure \ref{fig:ALL} (3-c) in the three tasks. From this search, we derive the Pareto Frontier, representing the set of optimal configurations that offer the best trade-offs between efficiency and effectiveness.



\begin{figure}[t] \centering
    \includegraphics[width=0.7\columnwidth]{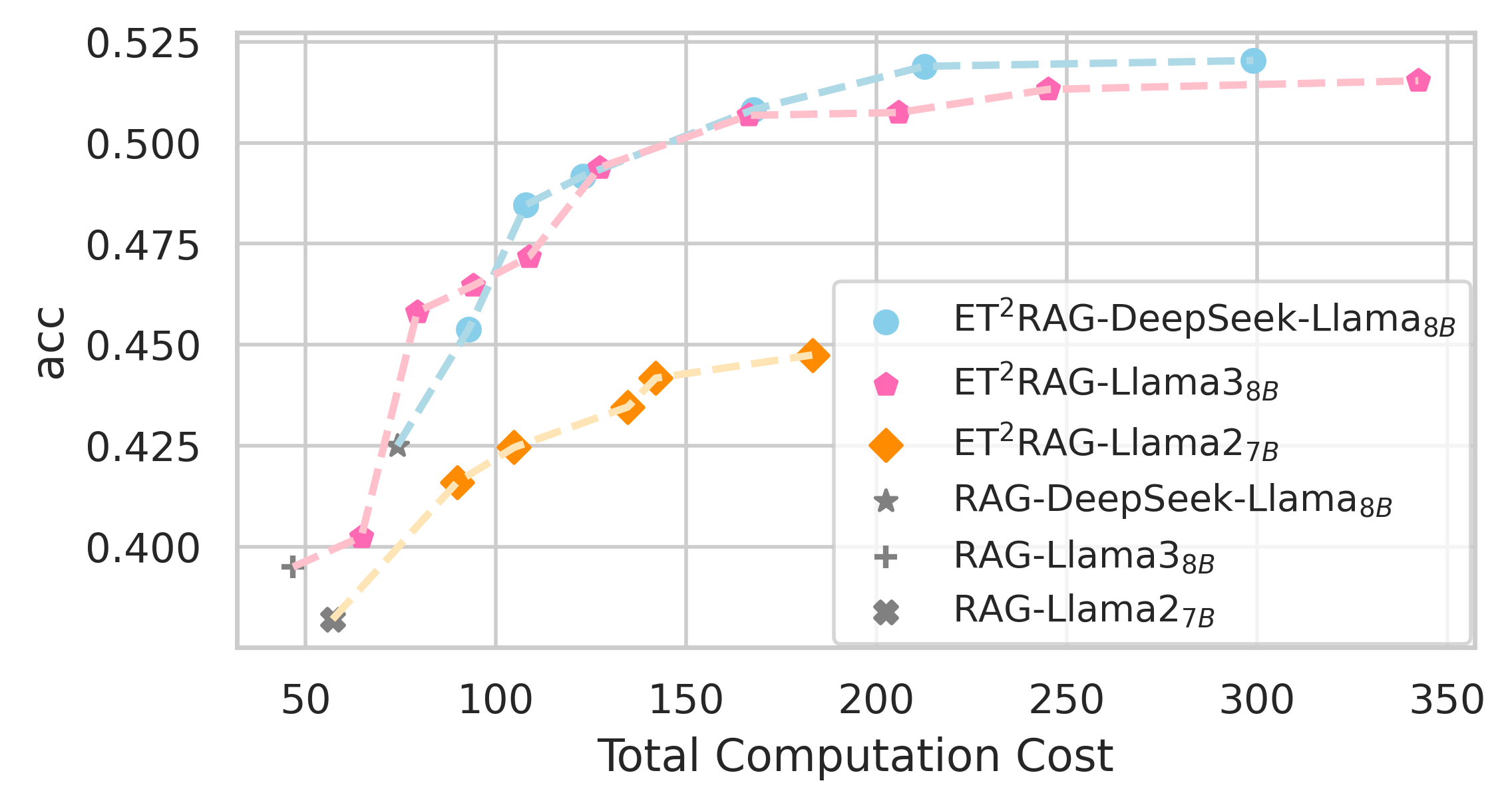}
    \caption{Trade-off between performance and total computation cost on PopQA.} \label{fig:pqa3}
\end{figure}

From Figure \ref{fig:ALL} (1-c), it is noticeable that for Open-domain question answering task in TriviaQA dataset, the combination of $V$ equal to $3$ and $L$ equal to $5$ offers the first cost-effective optimal solution. To enhance performance further, priority should be given to increasing $V$, followed by an increase in $L$. The improvements tend to decrease when $L$ escalates excessively. This suggests that generating complete answers for each candidate output set $O$ is not imperative in these tasks. By generating shorter responses, computational resources can be preserved while maintaining high-quality generation.
For the recipe generation task, as depicted in Figure \ref{fig:ALL} (2-c), the first cost-effective optimal solution is achieved when $V$ equals $3$ and $L$ equals $10$. 
If further enhancement in performance is desired, increasing either $V$ or $L$ yields similar gains. 
Lastly, for the image caption task in COCO dataset, as seen in Figure \ref{fig:ALL} (3-c), performance improvement requires $L$ to achieve a certain length. In this case, the first cost-effective optimal solution is achieved when $V$ equals $3$ and $L$ equals $3$. If further enhancement in performance is desired, $L$ should be increased first, followed by a rise in $V$.

Furthermore, taking PopQA as an example, we also calculate the total computational cost, plot the Pareto frontiers of different models, and include the original RAG method. Figure \ref{fig:pqa3} clearly shows the trade-off between computational cost and performance and the comparison to the RAG method. Notably, under the same computational budget, both DeepSeek-R1-Llama$_{8B}$~\cite{guo2025deepseek} and Llama3$_{8B}$~\cite{grattafiori2024llama} achieve better performance than Llama2$_{7B}$~\cite{touvron2023llama}, demonstrating the efficiency improvements of our method.

\subsection{Discussion} 
Our experimental analysis highlights that retrieval organization strategies must be tailored to the specific characteristics and demands of each task. 
In open-domain QA tasks, the combination strategy $\{\text{top}_1, \text{top}_k\}$ consistently outperforms the simpler $\{\text{top}_k\}$ approach. As shown in Figure~\ref{fig:abl}, relying solely on lower-ranked documents fails to yield improvements through majority voting. This suggests that, in QA settings, the highest-ranked document typically contains essential factual information, while additional documents may introduce marginally useful context but also increase the risk of injecting noise or redundancy.
For recipe generation tasks, our results show that the ${\text{top}_k}$ strategy alone is sufficiently effective. This observation is consistent with prior work~\cite{liu2024retrievalaugmentedrecipegeneration, salvador2021revamping}, which highlights that top-retrieved recipes often contain rich and informative content. The inherently structured and detailed nature of recipe texts means that even mid-ranked results typically offer relevant guidance. Moreover, limiting the input to top-$k$ results helps control the input length—an important consideration given the long cooking instructions, thereby reducing computational overhead without sacrificing output quality.
In image captioning tasks, we adopt the retrieval-combination scheme in \cite{ramos2023smallcap}, selecting diverse caption candidates by grouping four samples from the top 20 retrieved captions. This method maintains quality baselines while preserving necessary diversity, matching the creative nature of the task. Experimental results demonstrate that this balanced strategy achieves better performance without increasing computational overhead. Overall, these observations highlight the flexibility and extensibility of ET$^2$RAG. While a universal organization strategy may not exist, the modularity of our method provides the flexibility to adapt retrieval structures to the nuances of individual tasks, thereby maximizing performance while controlling efficiency.

\section{CONCLUSION}
We have presented an efficient and training-free method, ET$^2$TAG, which achieves state-of-the-art performance across three different tasks. A central insight of our work is that generating full-length responses in retrieval-augmented generation is often unnecessary; in fact, leveraging truncated outputs can lead to more efficient and even more accurate results by facilitating faster and more effective consensus. Our experiments demonstrate that the proposed organization strategy plays a critical role in maintaining the stability and relevance of retrieved information throughout the integration process. Additionally, we systematically analyzed the effects of key parameters, such as response length and voting size, revealing consistent trends that support the effectiveness of our design choices. These findings establish a solid foundation for future research on efficient integration of retrieval and generation in large language models.

\bibliographystyle{ACM-Reference-Format}
\bibliography{sample-base}

\appendix
\end{document}